\documentclass[]{spie}  

\usepackage{amsmath,amsfonts,amssymb}
\usepackage{graphicx}
\usepackage[colorlinks=true, allcolors=blue]{hyperref}

\title{BoxingVI: A Multi-Modal Benchmark for Boxing Action Recognition and Localization\\
\thanks{Supported by Ministry of Youth Affairs and Sports}
}

\author[a]{Rahul Kumar\textsuperscript{\dag}}
\author[a]{Vipul Baghel\textsuperscript{\dag}}
\author[a]{Sudhanshu Singh}
\author[a]{Bikash Kumar Badatya}
\author[c]{Shivam Yadav}
\author[d]{Babji Srinivasan}
\author[a]{Ravi Hegde}

\affil[a]{Department of Electrical Engineering, Indian Institute of Technology Gandhinagar, Gandhinagar, India}
\affil[c]{Department of Computer Science, Dr. A. P. J. Abdul Kalam Technical University, Uttar Pradesh, India}
\affil[d]{Applied Mechanics and Biomedical Engineering, Indian Institute of Technology Madras, Tamil Nadu, India}

\authorinfo{
Send correspondence to R.H.\\
R.K.: kmrrahul@iitgn.ac.in\\
V.B.: baghelvipul@iitgn.ac.in\\
S.S.: singhsudhanshu@iitgn.ac.in\\
B.K.B.: bikash.badatya@iitgn.ac.in\\
S.Y.: shivam007yadavji007@gmail.com\\
B.S.: babji.srinivasan@zmail.iitm.ac.in\\
R.H.: E-mail: hegder@iitgn.ac.in
}

\begin{document} 
\maketitle

\begin{abstract}
Accurate analysis of combat sports using computer vision has gained traction in recent years, yet the development of robust datasets remains a major bottleneck due to the dynamic, unstructured nature of actions and variations in recording environments. In this work, we present a comprehensive, well-annotated video dataset tailored for punch detection and classification in boxing. The dataset comprises $6,915$ high-quality punch clips categorized into six distinct punch types, extracted from $20$ publicly available YouTube sparring sessions and involving $18$ different athletes. Each clip is manually segmented and labeled to ensure precise temporal boundaries and class consistency, capturing a wide range of motion styles, camera angles, and athlete physiques. This dataset is specifically curated to support research in real-time vision-based action recognition, especially in low-resource and unconstrained environments. By providing a rich benchmark with diverse punch examples, this contribution aims to accelerate progress in movement analysis, automated coaching, and performance assessment within boxing and related domains.
\end{abstract}

\keywords{Sports Analytics, Human Pose Estimation, Human Pose Tracking, Human Action Recognition, Temporal Action Localization}

\section{INTRODUCTION}
\label{sec:intro}  

Sports analytics has revolutionized the way athletic performance is studied and optimized~\cite{cardenas2024beyond}. From strategy formulation to skill refinement, the use of data-driven methods has become integral across disciplines~\cite{shan2023data, barbon2024data}. While extensive datasets exist for mainstream sports like basketball~\cite{farghaly2024leveraging}, cricket~\cite{raajesh2024cricket}, tennis~\cite{sampaio2024applications}, and football~\cite{javed2023football}, combat sports remain underrepresented in publicly available, high-quality datasets—particularly for vision-based analysis.

\clearpage 
\begin{figure*}[p]
    \centering
    \includegraphics[width=\textwidth]{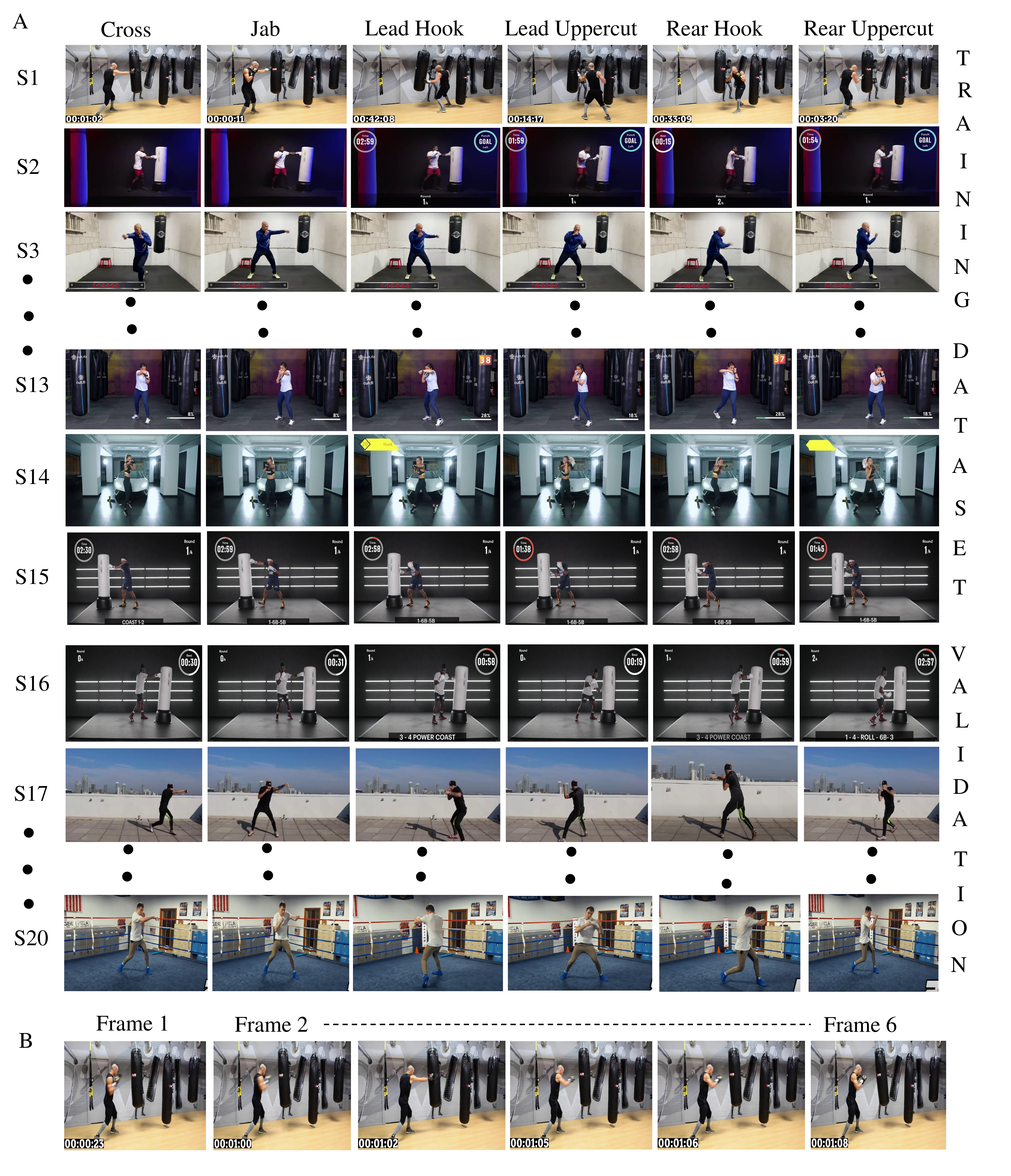}
    \caption{The dataset comprises $20$ different subjects from YouTube videos. \textbf{A}: Subjects S1 to S15 are used for training, while S16 to S20 are used for validation. Each column corresponds to a punch class (Cross, Jab, Lead Hook, Lead Uppercut, Rear Hook, Rear Uppercut) shown across subjects. \textbf{B}: A sequence from frame 1 to 6 depicts a cross punch from initiation to completion. \textit{Thumbnails are reproduced under the Fair Use Policy of YouTube.}}
    \label{fig:dataset}
\end{figure*}
\clearpage

An emerging focus within this domain is the use of computer vision to capture unstructured, fast-paced actions in sports such as boxing and Mixed Martial Arts (MMA)~\cite{chen2024interpretable}. While sensor-based methods—such as Inertial Measurement Units (IMUs)—have historically been used to capture movement patterns in combat sports~\cite{jayakumar2024multi, beranek2023force}, they pose several limitations: device drift, restricted movement, discomfort during usage, and risk of injury~\cite{khasanshin2021application, hanada2024boxing}. These factors hinder their deployment in realistic training and competition scenarios~\cite{gatt2023effects}.

To address these challenges, computer vision offers a non-intrusive, scalable alternative for motion analysis in combat sports~\cite{dos2021novel}. However, despite its potential, current approaches often rely on high-cost hardware or operate under controlled environments~\cite{host2022overview}. This restricts their accessibility for home-based or amateur use. Moreover, existing public datasets for combat sports lack sufficient granularity, diversity, and temporal annotations required for robust action recognition and segmentation.

In this context, we present a novel contribution: a large-scale, well-annotated dataset specifically designed for vision-based boxing analysis. The dataset comprises \textbf{6,915} finely segmented and labeled punch clips, categorized into six distinct punch types: \textit{Cross}, \textit{Jab}, \textit{Lead Hook}, \textit{Lead Uppercut}, \textit{Rear Hook}, and \textit{Rear Uppercut}. These clips are extracted from \textbf{20} publicly available YouTube sparring and air-boxing sessions, featuring \textbf{18} athletes (11 male, 7 female) across diverse styles, speeds, and settings. Key annotations include punch start/end time stamps and pose keypoints, providing both temporal and spatial context for each action. This work addresses a critical gap in the availability of datasets tailored to unstructured, real-world combat footage and lays the groundwork for downstream tasks such as action localization, technique classification, and skill assessment.

The rest of the paper is organized as follows:
Section \ref{sec:relatedwork} explains the review of the literature and the gaps present in the existing state-of-the-art. Section \ref{subsec:datasetcreation} describes the detailed methodology to create the proposed benchmark. Finally, we conclude our work in Section \ref{sec:conclusion} with its future aspects.  

\begin{figure*}[t!]
    \centering
    \includegraphics[width=1.0\textwidth]{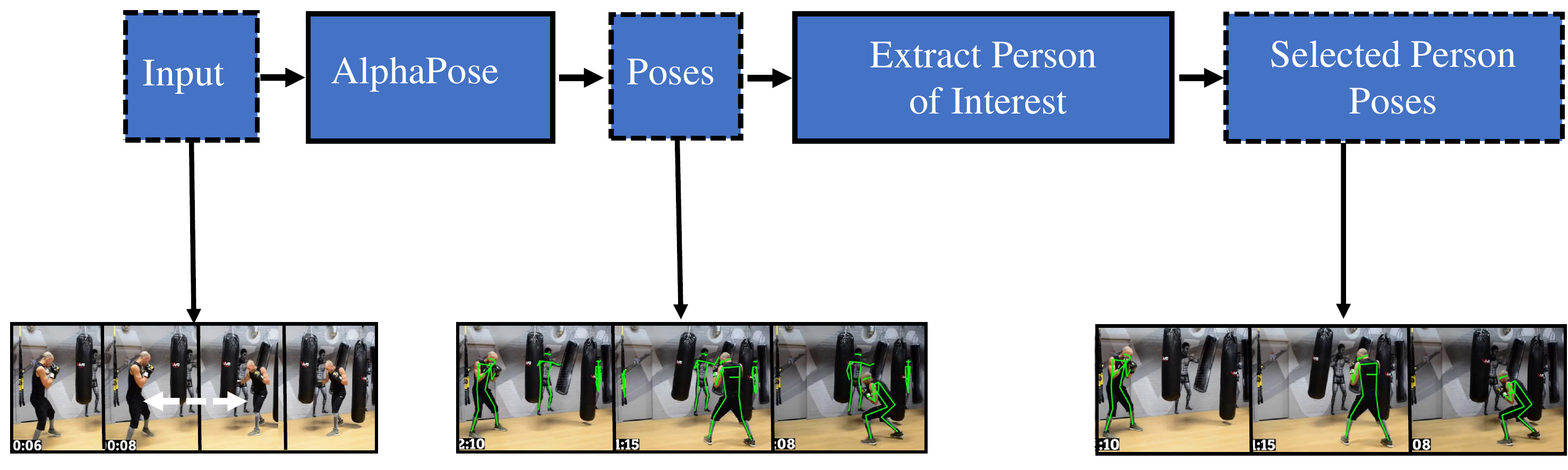}
    \caption{Tracking the person of interest across the video using the least Euclidean distance method applied to the detected poses. By determining the center of mass from the keypoints of the shoulders and hips, the individual is identified and consistently followed throughout the videos.  \textit{Thumbnails are reproduced under the Fair Use Policy of YouTube.}}
    \label{frame}
\end{figure*}

\section{RELATED WORK}
\label{sec:relatedwork}

Recent advances in computer vision and deep learning have significantly influenced the domain of sports analytics, particularly in tasks such as human action recognition (HAR), action localization, and temporal segmentation. These tasks are fundamentally dependent on the availability of high-quality annotated datasets that offer diverse modalities—typically RGB video, pose estimations, and frame-level action labels. However, while generic action recognition has benefited from large-scale datasets, there remains a notable gap in datasets tailored to the challenges of fine-grained, fast-paced combat sports like boxing.

Several large-scale datasets have been developed for generic action recognition in unconstrained environments. UCF101~\cite{soomro2012ucf101} and HMDB51~\cite{kuehne2011hmdb} are among the earliest benchmarks for action classification, composed of short video clips labeled with coarse activity categories. Though widely used for training and benchmarking deep video models, these datasets are limited by their trimmed format, lack of temporal boundaries, and absence of multi-modal annotations such as pose or depth. ActivityNet~\cite{caba2015activitynet} and Kinetics~\cite{kay2017kinetics} address some of these limitations by providing untrimmed video segments with temporal boundaries for action localization tasks. Kinetics, in particular, has been instrumental in training large transformer-based video models, owing to its large number of diverse activity classes. However, despite their scale and diversity, these datasets are not domain-specific and thus lack the granularity and semantic richness required for modeling structured actions in sports contexts.

Within sports analytics, some datasets have been proposed to address domain-specific requirements. The Sports-1M dataset~\cite{karpathy2014large} collects YouTube videos across a broad range of sports categories, but annotations are at the video level and lack precise temporal granularity. In the context of racket sports, Rahmad et al.~\cite{rahmad2020vision} introduced a badminton-specific dataset annotated for stroke classification, yet it lacks detailed temporal segmentation and pose information. Similarly, datasets for basketball or soccer, such as those used in tracking or tactical analysis~\cite{cuperman2022end}, focus on player trajectories or team dynamics rather than detailed action semantics. Stefański et al.~\cite{stefanski2023classification, stefanski2024boxing} proposed boxing-specific datasets with coarse-grained classification of events like punches and blocks, but their annotations are limited to entire rounds or predefined video segments, without frame-level temporal localization or fine-grained punch categorization. Moreover, most of these datasets rely solely on RGB video and do not incorporate skeletal representations, limiting their utility in structure-aware learning.

In parallel, skeleton-based datasets such as NTU RGB+D~\cite{shahroudy2016ntu} and NTU RGB+D 120~\cite{liu2019ntu120} have enabled progress in pose-based action recognition. These datasets offer high-resolution 3D joint coordinates and cover a wide range of daily activities, collected using Kinect sensors in controlled environments. While effective for general-purpose HAR, their applicability to combat sports is limited due to the lack of domain-specific actions, realistic motion patterns, and interaction dynamics present in sports like boxing. The Kinetics Skeleton dataset, derived from 2D pose estimators applied to Kinetics videos, allows pose-based modeling over large-scale RGB data, but inherits the limitations of the underlying pose estimator and lacks ground-truth skeletal annotations. The BABEL dataset~\cite{punnakkal2021babel} provides temporally dense frame-level annotations over motion capture recordings, offering a rich taxonomy of human actions in 3D. However, these recordings are scripted and lack alignment with visual content, making them unsuitable for training appearance-based models or studying real-world athletic motion.

More broadly, most existing datasets for action understanding are either too generic or too controlled, thereby failing to bridge the gap between visual realism, structural annotations, and semantic granularity. Combat sports like boxing pose unique challenges that are underrepresented in existing datasets: actions occur at high speed with frequent occlusions, involve multiple interacting participants, and require fine-grained distinction between visually similar movements such as a jab and a cross. Furthermore, monocular view constraints, variable lighting, and camera motion introduce significant noise, which is rarely modeled in benchmark datasets collected under laboratory settings.

In light of these limitations, there is a pressing need for datasets that are domain-specific to combat sports and that incorporate temporal, visual, and pose modalities in alignment. A dataset offering frame-level action demarcation, detailed punch-type annotations, 2D skeletal representations, and real-world monocular RGB videos would enable comprehensive benchmarking of models for action localization, structural modeling, and sequential prediction. Such a resource would also support practical downstream applications including automated scoring, athlete training feedback, and digital twin construction. Our dataset is designed precisely to fill this gap, offering a structured and fine-grained resource for the community working on real-world sports analytics and combat action understanding.

\section{Dataset Creation}{\label{subsec:datasetcreation}}
In this study, we introduce a boxing-specific dataset composed of 20 unedited YouTube videos featuring distinct individuals performing various boxing techniques, including shadow boxing, punching bag workouts, and instructional boxing bag tutorials. These videos collectively contain a wide variety of motion dynamics under real-world recording conditions. Subjects S1 to S15 are designated for the training set, while subjects S16 to S20 are reserved for validation. All punching sequences are recorded using a fixed monocular camera setup, as illustrated in Figure~\ref{frame}, ensuring consistency in viewpoint across the dataset. Two types of annotations are performed on this video corpus: (1) fine-grained temporal segmentation of punch clips with class-wise labeling for jab, cross, hook, and uppercut; and (2) frame-wise 2D human pose estimation focusing on the primary boxer in the scene. 

As summarized in Table~\ref{tab:dataset}, the proposed dataset contains a total of 6,915 temporally segmented clips (5,513 for training and 1,402 for validation), each annotated with its respective punch category and corresponding pose sequence. Compared to prior works, the 3DCG dataset~\cite{watanabepunch} provides 6,900 labeled clips derived from synthetic or edited recordings but lacks 2D pose annotations and temporal punch demarcation. Similarly, the BoxMAC dataset~\cite{sahoo2024boxmac}, though composed of 15 unedited videos yielding 2,314 clips (1,584 training and 730 testing), also lacks skeletal pose data and fine-grained action boundary annotations. In contrast, our dataset is the only one to provide raw RGB videos, temporal punch segmentation, per-clip class labels, and 2D human pose trajectories, thus supporting a broader range of tasks including temporal action localization, pose-conditioned recognition, and fine-grained action segmentation in combat sports scenarios.

By focusing on real-world, unconstrained video data captured from monocular views, our dataset fills a critical gap in current sports analytics resources, which predominantly rely on constrained laboratory settings, wearable IMU sensors, or synthetic data. This dataset enables diverse applications including temporal action localization, pose-conditioned action recognition, motion segmentation, and athlete performance profiling. Furthermore, it serves as a robust benchmark for evaluating the temporal consistency, spatial accuracy, and generalization capacity of deep learning models under realistic conditions.

\begin{table*}[ht]
\caption{Summary of publicly available boxing punch datasets. 
The table compares the Ours, 3DCG, and BoxMAC datasets, showing the number of samples for Clipped Video and the availability of annotations for RGB Video, Punch Classes, 2D Pose, and Punch Demarcation. 
This overview highlights the varying data availability and annotations across different datasets used for punching action recognition.}
\label{tab:dataset}
\centering
\footnotesize
\renewcommand{\arraystretch}{1.2}
\setlength{\tabcolsep}{2pt}
\begin{tabular}{lcccccccc}
\hline
\textbf{Dataset Name} & \textbf{Unedited Video} & \multicolumn{2}{c}{\textbf{Clipped Video}} & \textbf{RGB Video} & \textbf{Punch Classes} & \textbf{2D Pose} & \textbf{Punch Demarcation} \\
\cline{3-4}
& & Training & Testing & & & & \\
\hline
Ours   & 20   & 5513  & 1402  & \checkmark & \checkmark & \checkmark & \checkmark \\
3DCG\cite{watanabepunch}   & ---  & 6000  & 900   & \checkmark  & \checkmark & \texttimes & \texttimes \\
BoxMAC\cite{sahoo2024boxmac} & 15   & 1584  & 730   & \checkmark  & \checkmark & \texttimes & \texttimes \\
\hline
\end{tabular}
\end{table*}

\begin{figure}[t]
    \centering
    \includegraphics[width=0.45\textwidth]{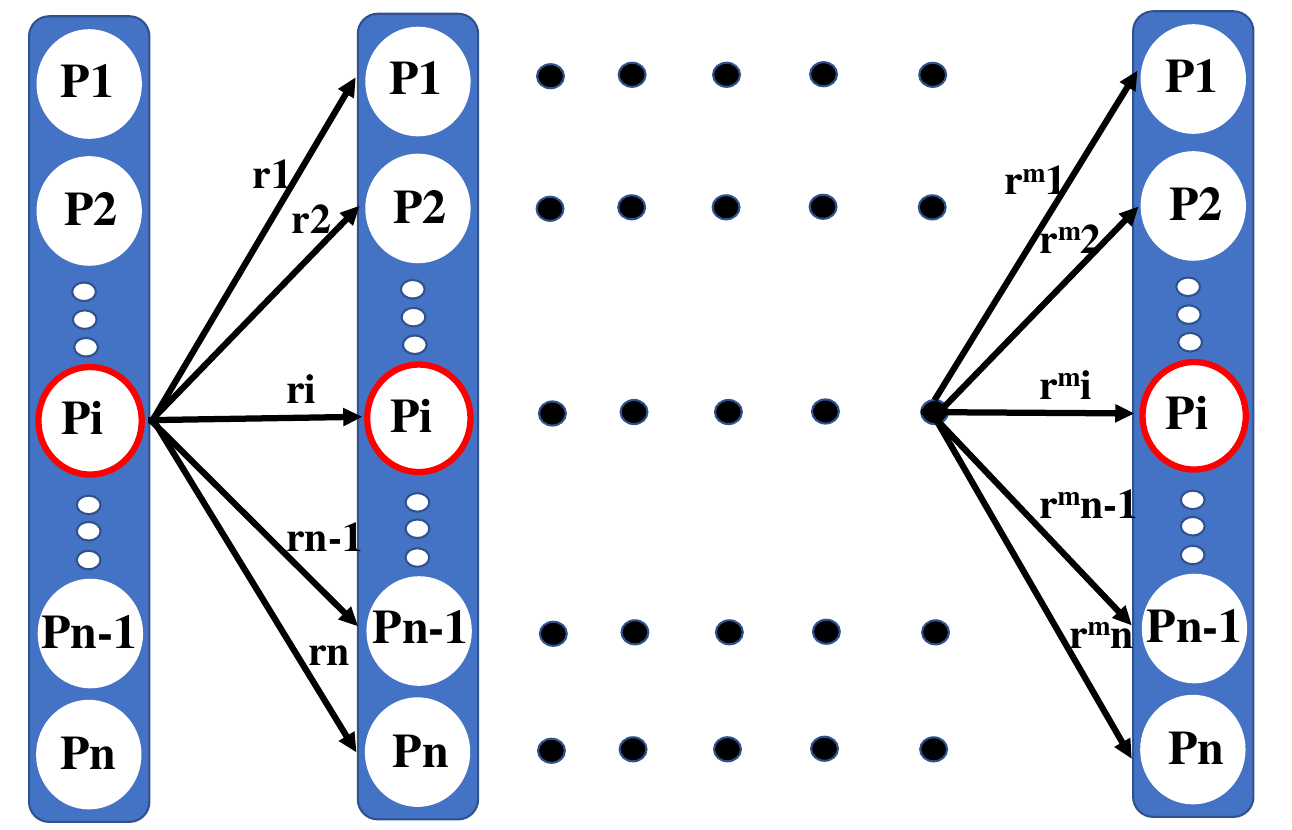}
    \caption{The least Euclidean distance method is used for tracking across \(m\) number of frames. \(P_1, P_2, \dots, P_n\) represent the total number of detected persons, where \(P_i\) is the person of interest. The Euclidean distance \(r_i\) is calculated between the position of the person of interest in the first frame and the \(i\)-th person in the second frame. The person with the least \(r_i\) is assumed to be the same as the person of interest.
}
    \label{frame1}
\end{figure}

\subsection{Pose Estimation}
For punch clip segmentation and precise boxing punch labeling, image frames are extracted from the 20 videos, and the start and end frames for each punch, along with the corresponding punch type, are manually documented for the person of interest. AlphaPose is employed to automatically detect and record human keypoints (poses), resulting in pose detection for multiple individuals in each video. To consistently identify the person of interest, a systematic approach is implemented. Each detected pose is localized by identifying the point closest to the center of mass, defined as the mean of both shoulders and hips. The person of interest is selected in the first frame and tracked across subsequent frames using the method of least displacement of the center of mass between consecutive frames. As illustrated in Figure~\ref{frame1}, multiple individuals (P1 to Pn) are detected in each frame. The person of interest (Pi) is manually selected in the first frame. Displacement of the center of mass for all individuals in the second frame relative to Pi in the first frame is calculated, denoted as r1 to rn. The individual with the least displacement is identified as the person of interest for the second frame, with this process repeated for each subsequent frame to ensure consistent tracking throughout the video.
\par
In the subsequent stage, the keypoints and corresponding frame numbers are extracted and normalized by dividing them by the video’s height and width to ensure consistent scaling, independent of the original video dimensions. Although the extracted pose sequences vary in length, statistical analysis of the dataset indicates that the maximum number of frames required to complete a punch in 30 fps videos is 25. To maintain temporal uniformity, sequences are padded with zeros up to 25 frames.

\subsection{Pose Tracking}
For the tracking methodology, multiple individuals (P1 to Pn) are detected in each frame, requiring a systematic approach to consistently identify the person of interest (Pi). Initially, Pi is manually selected in the first frame. To track Pi across subsequent frames, the displacement of the center of mass for all detected individuals is calculated relative to Pi in the previous frame. The individual with the least displacement is identified as Pi in the next frame, ensuring consistent tracking throughout the video. This process is repeated iteratively across the entire sequence. Figure~\ref{frame1} visually represents this tracking approach, showing how different individuals are identified frame by frame. The displacement values computed during tracking are denoted as r1 to rn, ensuring precise motion consistency throughout the sequence.

\section{Conclusion}
\label{sec:conclusion}

This work introduces a large-scale, well-annotated boxing punch dataset designed to advance computer vision research in combat sports. The data set comprises $6,915$ class-labeled and temporally segmented punch clips in six fine-grained categories, derived from $20$ publicly available YouTube videos featuring $18$ diverse athletes involved in shadow boxing, bag training, and sparring drills. Each punch clip is annotated with precise temporal boundaries and 2D pose keypoints, facilitating fine-grained analysis of motion execution and inter-frame dynamics.

Beyond its utility in computer vision research, this dataset holds promise for applied domains such as automated coaching systems, intelligent broadcast analytics, digital twin construction, sports biomechanics, and rehabilitation monitoring. Future work will focus on expanding the dataset to incorporate multi-person combat scenarios, opponent interactions, and additional sports disciplines to enhance its generalizability. Such extensions will facilitate the development of next-generation AI models for real-time skill assessment, strategic behavior modeling, and personalized feedback in dynamic, multi-agent environments.

\subsection*{Availability of Data}
The dataset include YouTube video links, temporal demarcation of punches and punch category labels. It doesn't directly include any copyrighted material. Dataset repository can be accessed via \url{https://github.com/Bikudebug/BoxingVI.git}.

\acknowledgments 
 
We acknowledge our collaborator boxing coach John Warburton from Inspire Inspire Institute of Sports. We also acknowledge YouTube to allow us to utilize their videos for our research under the Fair Use Policy.

\bibliography{report} 

@article{khasanshin2021application,
  title={Application of an artificial neural network to automate the measurement of kinematic characteristics of punches in boxing},
  author={Khasanshin, Ilshat},
  journal={Applied Sciences},
  volume={11},
  number={3},
  pages={1223},
  year={2021},
  publisher={MDPI}
}

@phdthesis{gatt2023effects,
  title={Effects of bandaging techniques and shot types on wrist motion in boxing},
  author={Gatt, Ian},
  year={2023},
  school={Sheffield Hallam University}
}

@article{dos2021novel,
  title={A novel feature extractor for human action recognition in visual question answering},
  author={dos S Silva, Francisco H and Bezerra, Gabriel M and Holanda, Gabriel B and de Souza, J Wellington M and Rego, Paulo AL and Neto, Alo{\'\i}sio V Lira and de Albuquerque, Victor Hugo C and Rebou{\c{c}}as Filho, Pedro P},
  journal={Pattern Recognition Letters},
  volume={147},
  pages={41--47},
  year={2021},
  publisher={Elsevier}
}

@article{jayakumar2024multi,
  title={Multi-sensor fusion based optimized deep convolutional neural network for boxing punch activity recognition},
  author={Jayakumar, Brindha and Govindarajan, Nallavan},
  journal={Proceedings of the Institution of Mechanical Engineers, Part P: Journal of Sports Engineering and Technology},
  pages={17543371241237085},
  year={2024},
  publisher={SAGE Publications Sage UK: London, England}
}

@inproceedings{rahmad2020vision,
  title={Vision based automated badminton action recognition using the new local convolutional neural network extractor},
  author={Rahmad, Nur Azmina and As’ ari, Muhammad Amir and Ibrahim, Mohamad Fauzi and Sufri, Nur Anis Jasmin and Rangasamy, Keerthana},
  booktitle={Enhancing Health and Sports Performance by Design: Proceedings of the 2019 Movement, Health \& Exercise (MoHE) and International Sports Science Conference (ISSC)},
  pages={290--298},
  year={2020},
  organization={Springer}
}

@article{sahoo2024boxmac,
  title={BoxMAC--A Boxing Dataset for Multi-label Action Classification},
  author={Sahoo, Shashikanta},
  journal={arXiv preprint arXiv:2412.18204},
  year={2024}
}

@article{watanabepunch,
  title={Punch Type Classification and Hit Judgement Using Estimated Skeletal Model in Boxing Match Videos},
  author={Watanabe, Soma and Kameda, Yoshinari},
  year={2024}
}

@article{cuperman2022end,
  title={An end-to-end deep learning pipeline for football activity recognition based on wearable acceleration sensors},
  author={Cuperman, Rafael and Jansen, Kaspar MB and Ciszewski, Micha{\l} G},
  journal={Sensors},
  volume={22},
  number={4},
  pages={1347},
  year={2022},
  publisher={MDPI}
}

@inproceedings{stefanski2023classification,
  title={Classification of Punches in Olympic Boxing Using Static RGB Cameras},
  author={Stefa{\'n}ski, Piotr and Jach, Tomasz and Kozak, Jan},
  booktitle={International Conference on Computational Collective Intelligence},
  pages={540--551},
  year={2023},
  organization={Springer}
}

@article{stefanski2024boxing,
  title={Boxing Punch Detection with Single Static Camera},
  author={Stefa{\'n}ski, Piotr and Kozak, Jan and Jach, Tomasz},
  journal={Entropy},
  volume={26},
  number={8},
  year={2024},
  publisher={Multidisciplinary Digital Publishing Institute (MDPI)}
}

@article{cardenas2024beyond,
  title={Beyond hard workout: A multimodal framework for personalised running training with immersive technologies},
  author={Cardenas Hernandez, Fernando Pedro and Schneider, Jan and Di Mitri, Daniele and Jivet, Ioana and Drachsler, Hendrik},
  journal={British Journal of Educational Technology},
  year={2024},
  publisher={Wiley Online Library}
}

@incollection{barbon2024data,
  title={Data-Driven Methods for Soccer Analysis},
  author={Barbon Junior, Sylvio and Moura, Felipe Arruda and da Silva Torres, Ricardo},
  booktitle={Artificial Intelligence in Sports, Movement, and Health},
  pages={233--253},
  year={2024},
  publisher={Springer}
}

@article{shan2023data,
  title={Data driven intelligent action recognition and correction in sports training and teaching},
  author={Shan, Sicong and Sun, Shuang and Dong, Peng},
  journal={Evolutionary Intelligence},
  volume={16},
  number={5},
  pages={1679--1687},
  year={2023},
  publisher={Springer}
}

@inproceedings{farghaly2024leveraging,
  title={Leveraging Machine Learning to Predict National Basketball Association Player Injuries},
  author={Farghaly, Omar and Deshpande, Priya},
  booktitle={2024 IEEE International Workshop on Sport, Technology and Research (STAR)},
  pages={216--221},
  year={2024},
  organization={IEEE}
}

@inproceedings{raajesh2024cricket,
  title={Cricket Team Selection and Player Analysis using Data Analytics},
  author={Raajesh, Sanjay and Martin, Noel and Jiji, Jyothsna and Nair, Aakash and Haritha, H},
  booktitle={2024 IEEE Recent Advances in Intelligent Computational Systems (RAICS)},
  pages={1--6},
  year={2024},
  organization={IEEE}
}

@article{sampaio2024applications,
  title={Applications of Machine Learning to Optimize Tennis Performance: A Systematic Review},
  author={Sampaio, Tatiana and Oliveira, Jo{\~a}o P and Marinho, Daniel A and Neiva, Henrique P and Morais, Jorge E},
  journal={Applied Sciences},
  volume={14},
  number={13},
  pages={5517},
  year={2024},
  publisher={MDPI}
}

@inproceedings{javed2023football,
  title={Football analytics for goal prediction to assess player performance},
  author={Javed, Danish and Jhanjhi, NZ and Khan, Navid Ali},
  booktitle={Innovation and Technology in Sports: Proceedings of the International Conference on Innovation and Technology in Sports,(ICITS) 2022, Malaysia},
  pages={245--257},
  year={2023},
  organization={Springer}
}

@article{chen2024interpretable,
  title={An interpretable composite CNN and GRU for fine-grained martial arts motion modeling using big data analytics and machine learning},
  author={Chen, Gang},
  journal={Soft Computing},
  volume={28},
  number={3},
  pages={2223--2243},
  year={2024},
  publisher={Springer}
}

@article{beranek2023force,
  title={Force and velocity of impact during upper limb strikes in combat sports: a systematic review and meta-analysis},
  author={Ber{\'a}nek, V{\'a}clav and Vot{\'a}pek, Petr and Stastny, Petr},
  journal={Sports biomechanics},
  volume={22},
  number={8},
  pages={921--939},
  year={2023},
  publisher={Taylor \& Francis}
}

@incollection{hanada2024boxing,
  title={Boxing Movements Recognition Using IMUs During Shadow Boxing Exercise},
  author={Hanada, Yoshinori and Hossain, Tahera and Yokokubo, Anna and Lopez, Guillaume},
  booktitle={Human Activity and Behavior Analysis},
  pages={232--248},
  year={2024},
  publisher={CRC Press}
}

@article{host2022overview,
  title={An overview of Human Action Recognition in sports based on Computer Vision},
  author={Host, Kristina and Iva{\v{s}}i{\'c}-Kos, Marina},
  journal={Heliyon},
  volume={8},
  number={6},
  year={2022},
  publisher={Elsevier}
}

@article{soomro2012ucf101,
  title={UCF101: A dataset of 101 human actions classes from videos in the wild},
  author={Soomro, Khurram and Zamir, Amir Roshan and Shah, Mubarak},
  journal={arXiv preprint arXiv:1212.0402},
  year={2012}
}

@inproceedings{kuehne2011hmdb,
  title={HMDB: A large video database for human motion recognition},
  author={Kuehne, Hildegard and Jhuang, Hueihan and Garrote, Esteban and Poggio, Tomaso and Serre, Thomas},
  booktitle={ICCV},
  pages={2556--2563},
  year={2011}
}

@inproceedings{caba2015activitynet,
  title={ActivityNet: A large-scale video benchmark for human activity understanding},
  author={Caba Heilbron, Fabian and Escorcia, Victor and Ghanem, Bernard and Niebles, Juan Carlos},
  booktitle={CVPR},
  pages={961--970},
  year={2015}
}

@inproceedings{kay2017kinetics,
  title={The Kinetics Human Action Video Dataset},
  author={Kay, Will et al.},
  journal={arXiv preprint arXiv:1705.06950},
  year={2017}
}

@inproceedings{karpathy2014large,
  title={Large-scale video classification with convolutional neural networks},
  author={Karpathy, Andrej and Toderici, George and Shetty, Sanketh and Leung, Thomas and Sukthankar, Rahul and Fei-Fei, Li},
  booktitle={CVPR},
  pages={1725--1732},
  year={2014}
}

@inproceedings{shahroudy2016ntu,
  title={NTU RGB+D: A large scale dataset for 3D human activity analysis},
  author={Shahroudy, Amir and Liu, Jun and Ng, Tian-Tsong and Wang, Gang},
  booktitle={CVPR},
  pages={1010--1019},
  year={2016}
}

@article{liu2019ntu120,
  title={NTU RGB+D 120: A large-scale benchmark for 3D human activity understanding},
  author={Liu, Jun and Shahroudy, Amir and Perez, Alejandro and Wang, Gang and Duan, Ling-Yu and Kot, Alex C},
  journal={TPAMI},
  volume={42},
  number={10},
  pages={2684--2701},
  year={2019}
}

@inproceedings{punnakkal2021babel,
  title={BABEL: Bodies, Action and Behavior with English Labels},
  author={Punnakkal, Aayush and Ci, Heng and Gall, J{\"u}rgen and Joo, Hanbyul and Pons-Moll, Gerard},
  booktitle={CVPR},
  pages={722--731},
  year={2021}
}
\bibliographystyle{spiebib} 

\end{document}